\documentclass{article}

\usepackage[nonatbib,preprint]{neurips_2022}

\usepackage[utf8]{inputenc} % allow utf-8 input
\usepackage[T1]{fontenc}    % use 8-bit T1 fonts
\usepackage{hyperref}       % hyperlinks
\usepackage{url}            % simple URL typesetting
\usepackage{booktabs}       % professional-quality tables
\usepackage{amsfonts}       % blackboard math symbols
\usepackage{nicefrac}       % compact symbols for 1/2, etc.
\usepackage{microtype}      % microtypography
\usepackage{xcolor}         % colors
\usepackage[toc,page]{appendix}
\usepackage{graphicx}
\usepackage{subfigure}
\usepackage{amsmath}
\usepackage{color,colortbl}
\usepackage{array}
\usepackage{multirow}
\usepackage{mathtools}
\usepackage{enumitem}
\usepackage{subfiles} % Best loaded last in the preamble

\title{FreDo: Frequency Domain-based Long-Term Time Series Forecasting}

\author{%
  Fan-Keng Sun \\
  Department of EECS, MIT \\
  \texttt{fankeng@mit.edu} 
  \And
  Duane S. Boning \\
  Department of EECS, MIT \\
  \texttt{boning@mtl.mit.edu}
}

\begin{document}

\maketitle

\vspace{-10pt}

\begin{abstract}
The ability to forecast far into the future is highly beneficial to many applications, including but not limited to climatology, energy consumption, and logistics.
However, due to noise or measurement error, it is questionable how far into the future one can reasonably predict.
In this paper, we first mathematically show that due to error accumulation, sophisticated models might not outperform baseline models for long-term forecasting.
To demonstrate, we show that a non-parametric baseline model based on periodicity can actually achieve comparable performance to a state-of-the-art Transformer-based model on various datasets.
We further propose \textsf{FreDo}, a frequency domain-based neural network model that is built on top of the baseline model to enhance its performance and which greatly outperforms the state-of-the-art model.
Finally, we validate that the frequency domain is indeed better by comparing univariate models trained in the frequency v.s. time domain.
% Finally, we analyze results and find that the frequency domain enables sparser representation, making these models easier to learn.
\end{abstract}

\vspace{-25pt}

\section{Introduction} \label{sec:intro}

\vspace{-10pt}

Time series forecasting is an interdisciplinary field that has a wide range of applications.
Scientists at weather stations research models that help predict tomorrow's temperature and precipitation.
Engineers at manufacturing plants study models that can anticipate malfunctions of machines.
Analysts at companies develop models that estimate the price and demand of products in the coming weeks.
All of these work in different fields with different kinds of data using different types of models, but all are deeply engaged in time series forecasting.

Most previous work on time series forecasting focuses on short-term forecasting, especially for the immediate $t+1$ term.
The terms that are closer to the present are easier to forecast and usually have higher importance in most cases.
However, the ability to perform long-term forecasting accurately is desirable.
With the exponentially increasing amount of data and computing power, combined with the advancements in deep learning, forecasting longer into the future might eventually be feasible.

However, intuitively, errors accumulate as we forecast farther and farther into the future.
Thus, it is questionable how far into the future one can accurately predict.
In fact, as shown in Sec.~\ref{sec:erracc}, a complex model might not outperform a simple baseline model that always outputs the average value as the horizon increases.
Various previous works~\cite{autocorrelation,informer} have ignored the problem of error accumulation and designed sophisticated models based on Transformers~\cite{Transformer} for long-term forecasting.
In this paper, we show that a baseline model based on periodicity actually achieves comparable results to such Transformer-based models on most datasets.
Additionally, we propose \textsf{FreDo}, a frequency domain-based model that is built on top of the baseline model to enhance its performance.
Finally, by univariately evaluating models based on the frequency v.s. time domain, we validate that learning in the frequency domain further improves model performance.
% we analyze the advantage of frequency domain vs. time domain based models, to understand the source of their advantage.

Our main contributions are:

\begin{itemize}
    \item We mathematically derive that, under some general constraints, errors accumulate in forecasting when we forecast farther into the future. We also show that complex models might not outperform a simple baseline model.
    \item We design \textsf{AverageTile}, a baseline model based on periodicity that performs comparably to \textsf{Autoformer}~\cite{autocorrelation}, the state-of-the-art Transformer-based time series model.
    \item We propose \textsf{FreDo} that is built on top of \textsf{AverageTile} which learns in the frequency domain and greatly outperforms \textsf{Autoformer}.
    \item We validate the advantage of frequency domain learning by comparing univariate models trained in the frequency v.s. time domain.
\end{itemize}

\vspace{-10pt}

\section{Related Work} \label{sec:related}

\vspace{-5pt}

Most time series forecasting models focus on predicting the next time step.
The most well-known are linear models, including the autoregressive (AR) model, the autoregressive integrated moving average (ARIMA) model, and the vector autoregression (VAR) model~\cite{BJ}.
For nonlinear data, before the widespread application of neural networks (NNs), typical approaches have also included kernel methods~\cite{kernel-method}, ensembles~\cite{ensemble-method}, Gaussian processes~\cite{Gaussian-Process_0}, and regime switching~\cite{SETAR}, which all apply some fixed sets of nonlinearities that may be limited or not suitable for some real-world datasets.
In contrast, NNs can learn from the data in a more flexible manner and thus have recently become prominent for time series forecasting.

Long Short-term Memory (LSTM)~\cite{LSTM} networks have been the most popular and basic time series forecasting model.
These are recurrent neural networks (RNNs) that generate predictions recurrently, and are building blocks for more complex NNs, such as DeepAR~\cite{deepar} and DA-RNN~\cite{da-rnn}.
It is also common to add convolutional neural network (CNN) components to the model to improve pattern recognition.
These include Temporal Convolutional Networks~\cite{TCN}, Temporal Pattern Attention~\cite{TPA}, and Dual Self-Attention Networks~\cite{DSANet}.
For time series on a graph, Adaptive Graph Convolutional Recurrent
Networks~\cite{AGCRN} aim to learn both temporal and spatial patterns by combining RNNs and graph neural networks (GNNs).
These are just just a few examples from the universe of NNs for time series forecasting.

Given substantial progress on next-step forecasting, recent efforts have sought to forecast further into the future.
A direct method to do so is to recurrently predict the next-time step using one of the aforementioned models.
However, this is not only time consuming, but also makes the model prone to error accumulation.
Observing the great success of Transformers~\cite{Transformer} in natural language processing for generating very long sentences, several works have designed sophisticated models based on Transformers for long-term time series forecasting.
LogTrans~\cite{logtrans} is the first to show that a Transformer-based model can perform well on time series forecasting by introducing local convolutions and LogSpare attention.
The Informer~\cite{informer} further targets long-term forecasting by designing ProbSparse attention with lower complexity.
Lastly, the Autoformer~\cite{autoformer} proposes the Auto-Correlation mechanism to capture series-wise periodic connections, and thus greatly outperforms both LogTrans and Informer in long-term forecasting.
However, our work shows that \textsf{FreDo}, a simple feed-forward NN in the frequency domain, can greatly outperform \textsf{Autoformer} by taking advantage of the periodicities in the data.

\vspace{-5pt}

\section{Preliminaries} \label{sec:pre}

\vspace{-5pt}

A time series dataset is collected by $N$ sensors that record signals starting from the same time with the same sample rate over $T$ time steps.
Consequently, the whole dataset is a matrix $\mathbf{X} = \{\mathbf{x}_1, \dots, \mathbf{x}_t, \dots, \mathbf{x}_T\} \in \mathbb{R}^{T \times N}$, where $\mathbf{x}_t = [x_{t,1}, \dots, x_{t,N}] \in \mathbb{R}^N$ are the recorded signals by all $N$ sensors at the $t$-th time step.
% Note that we use normal lowercase letters, bold lowercased letters, and bold uppercased letters to denote scalars, vectors, and matrices respectively.

The goal of forecasting is to predict the future horizons $\{\mathbf{x}_t, \mathbf{x}_{t+1}, \dots\}$ given the histories $\{\mathbf{x}_1, \dots, \mathbf{x}_{t-1}\}$.
We use $\{\mathbf{\hat{x}}_{t}, \mathbf{\hat{x}}_{t+1}, \dots\}$ to denote the  predictions made by a model.
In practice~\cite{LSTNet,autoformer,MTGNN,informer}, only the $I$ most recent histories $\{\mathbf{x}_{t-I}, \dots, \mathbf{x}_{t-1}\}$ are fed into a model as input.
This not only establishes fair comparisons between different methods, but also makes the memory usage reasonable while assuming histories before $t-I$ have little influence on the future.

Currently, to the best of our knowledge, there is no precise definition as to what is considered short-term or long-term in time series forecasting.
In general, as stated in~\cite{informer}, the purpose of long-term forecasting is to forecast significantly longer into the future than most previous methods can achieve.
Additionally, following the description and setup in~\cite{autoformer}, we can specify long-term forecasting as predicting $O$ future horizons given $I$ histories where $O \geq I$, although the value of $I$ is actually a choice of hyperparameter rather than a part of the problem statement.
Combining both statements, we consider long-term forecasting to be when $O \geq I$ and both $I$ and $O$ are much larger than those achieved in most previous work.

\vspace{-5pt}
\section{Error Accumulation in Long-term Forecasting} \label{sec:erracc}
\vspace{-5pt}
Errors or noises in time series are almost unavoidable.
In terms of forecasting, we have the following data-generating process (DGP):
\vspace{-11pt}
\begin{equation}
    x_t = f(x_{t-1}, x_{t-2}, \dots) + e_t, 
    \label{eq:dgp}
\end{equation}
where $f$ can be any function and $e_t$ is the error with mean $0$ and variance $\sigma^2$.
Here, $e_t$ at different time steps are assumed to be mutually independent, though in the real world this might not be the case~\cite{autocorrelation}.
Furthermore, imagine that we already know the exact parameters of the function $f$ in Eq.~(\ref{eq:dgp}), which is nearly impossible in practice.

Usually, the scale of $e_t$ is smaller than $f(x_{t-1}, x_{t-2}, \dots)$ so the prediction (under DGP) is useful.
However, errors accumulate over time.
Hence, conceptually, the errors of model predictions increase as we try to forecast farther and farther into the future.
Eventually, the error is so large that it renders the model useless even if we know the DGP.

To formally show error accumulation, we assume that the DGP is a $p$-th order autoregressive (AR) model:
\vspace{-14pt}
\begin{equation}
    x_t = c + \sum_{i=1}^p \theta_p x_{t-p} + e_t = c + (\sum_{i=1}^p \theta_p L^{i-1}) x_{t-p} + e_t \coloneqq c + \phi(L; \theta) x_{t-1} + e_t,
    \label{eq:ardgp}
\end{equation}
where $\theta_1, \dots, \theta_p$ are the parameters of the DGP, $c$ is a constant, $L$ is the lag operator, and $\phi(L; \theta) = \sum_{i=1}^p \theta_p L^{i-1}$ is a polynomial of lag operators.
Now, after observing $p$ data points $x_0, \dots, x_{p-1}$ as input and assuming $x_t = 0, \forall t < 0$, we want to forecast $x_p, x_{p+1}, \dots$ using the DGP.
Notice that $\text{Var}[x_t] = \text{Var}[e_t] = 0, \forall t < p$ because $x_0, \dots, x_{p-1}$ are observed values, but $\text{Var}[e_t] = \sigma^2, \forall t \geq p$.
To illustrate, for the first two forecast steps, we have
\vspace{-6pt}
\begin{align}
    x_p &= c + \phi(L; \theta)x_{p-1} + e_p \text{ and}\\
    x_{p+1} &= c + \phi(L; \theta)x_{p} + e_{p+1} \label{eq:rep0} \\
            &= c + \phi(L; \theta)(c + \phi(L; \theta) x_{p-1} + e_p) + e_{p+1} \label{eq:rep1} \\
            &= (1 + \phi(L; \theta)) c + \phi^2(L; \theta) x_{p-1} + \phi(L; \theta) e_p + e_{p+1}. \label{eq:expand}
\end{align}
Thus, we have
\vspace{-6pt}
\begin{align*}
    \text{Var}[x_p] &= \text{Var}[e_p] = \sigma^2 \text{ and} \\[-8pt]
    \text{Var}[x_{p+1}] &= \text{Var}[\phi(L; \theta) e_p + e_{p+1}] = \text{Var}[\sum_{i=1}^p \theta_i e_{p+1-i} + e_{p+1}] = (\theta^2_1 + 1) \sigma^2. \\[-17pt]
\end{align*}
We can see that the variance increases from time step $p$ to $p+1$, which means that the best mean-squared error (MSE) achievable by any model increases strictly monotonically.
Next, we want to show that this is true for any $t \geq p$.
Following Eq.~(\ref{eq:ardgp}), we can always replace the $x$ terms on the right-hand-side with greater lag terms, for example from Eq.~(\ref{eq:rep0}) to Eq.~(\ref{eq:rep1}).
Eventually, similar to Eq.~(\ref{eq:expand}), we have
\vspace{-6pt}
\begin{equation}
    x_{p+k} = c \sum_{i=0}^k \phi^i(L; \theta) + \phi^{k+1}(L; \theta) x_{p-1} + \sum_{i=0}^k \phi^i e_{p+k-i}, \forall k \geq 0.
\end{equation}
Then, its variance is $\text{Var}[x_{p+k}] = \text{Var}[\sum_{i=0}^k \phi^i e_{p+k-i}]$, because the constant term and the $x_{p-1}$ term have zero variance.
If we compare the variance of $x_{p+k}$ and $x_{p+k+1}$:
\begin{align*}
    \text{Var}[x_{p+k}] &= \text{Var}[\sum_{i=0}^k \phi^i e_{p+k-i}] = \text{Var}[e_{p+k} + \phi(L; \theta) e_{p+k-1} + \phi^2(L; \theta) e_{p+k-2} + \dots], \\
    \text{Var}[x_{p+k+1}] &= \text{Var}[\sum_{i=0}^{k+1} \phi^i e_{p+k+1-i}] = \text{Var}[e_{p+k+1} + \phi(L; \theta) e_{p+k} + \phi^2(L; \theta) e_{p+k-1} + \dots],
\end{align*}
we can see that the coefficients for the error terms are the same for the first $k$ terms, but the variance of $x_{p+k+1}$ has one additional $k+1$-th term.
That is, if we expand all lag polynomials, group coefficients by different error terms, and sum the variances, we can conclude that $\text{Var}[x_{p+k+1}] > \text{Var}[x_{p+k}]$.
This means that the best achievable MSE increases strictly monotonically no matter what model is used, even for non-recurrent, multi-horizon prediction models.

The above derivation applies to AR models.
For general nonlinear function $f$, the best we can do is to use first-order Taylor expansion, so the same applies.
If the time series is not stationary, the variance can increase to infinity, which renders any model, including the DGP, useless in the long-term.
In the case of weakly stationary time series, it is assumed that both $\mathbb{E}[x_t]$ and $\text{Var}[x_t]$ are fixed values as $t \rightarrow \infty$.
Thus, there is a bound on the variance, which is exactly $\text{Var}[x_t]$.
In other words, the MSE starts from $\sigma^2$ at the first step and approaches $\text{Var}[x_t]$ strictly monotonically as we forecast farther into the future.
Although the variance is bounded, the optimal model prediction is still $\mathbb{E}[x_t]$.
This, again, means that any model is not better than just predicting the average of observed data points.

Finally, we want to point out that Eq.~(\ref{eq:dgp}) means that the value of $x_t$ is neither dependent on other exogenous variables nor on the timestamp $t$. If this is not the case, then it is possible that the error does not accumulate at all.
For instance, if the DGP is instead $x_t = f(t) + e_t$, then there is no error accumulation.
In the real world, the data is probably a mix of $f(x_{t-1}, \dots)$ and $f(t)$, so the error accumulation might not increase monotonically but will still trend upward over the long-term.

Notice that the model is still useful if the forecast horizon is short.
There is no clear cutoff with respect to horizon as to whether a model is useful vs. useless, especially for real-world datasets.
However, we emphasize that in many cases a sophisticated model might not outperform a simple baseline model (e.g., always predicting the average value) when the forecast horizon is long enough.
Indeed, in the next section, we propose a baseline model, which astonishingly achieves similar performance to the state-of-the-art Transformer-based \textsf{Autoformer} model~\cite{autoformer} on most datasets considered.

\begin{figure}[t!]
\centering

\begin{minipage}{0.48\textwidth}
\centering
\includegraphics[width=\columnwidth]{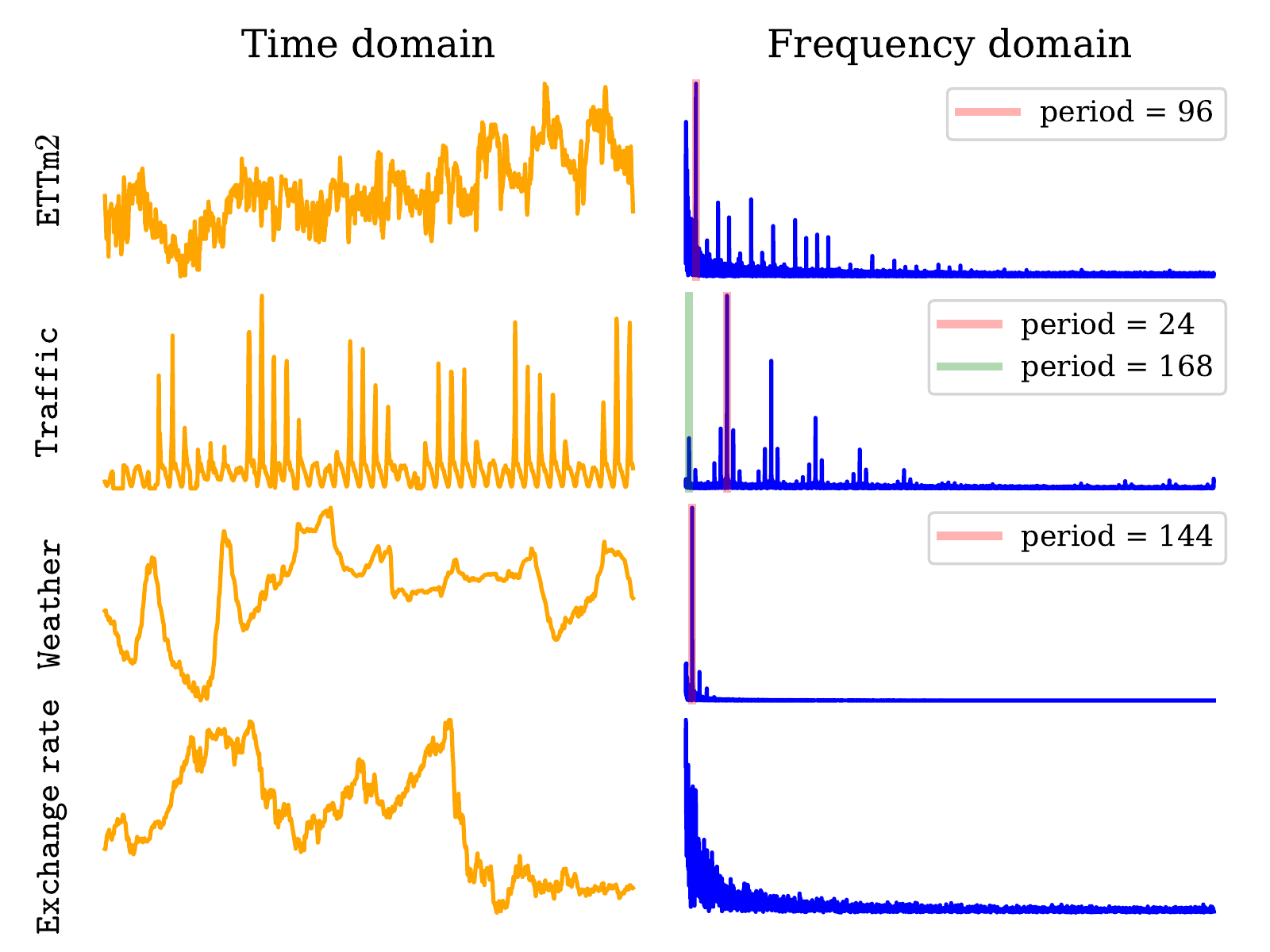}
\caption{Periodicities observed in time series datasets. Among four datasets shown here, only  \texttt{Exchange} has no meaningful periodicity.}
\label{fig:period_obs}
\end{minipage} %
\begin{minipage}{0.02\textwidth}
~
\end{minipage}
\begin{minipage}{0.45\textwidth}
\centering
\includegraphics[width=\columnwidth]{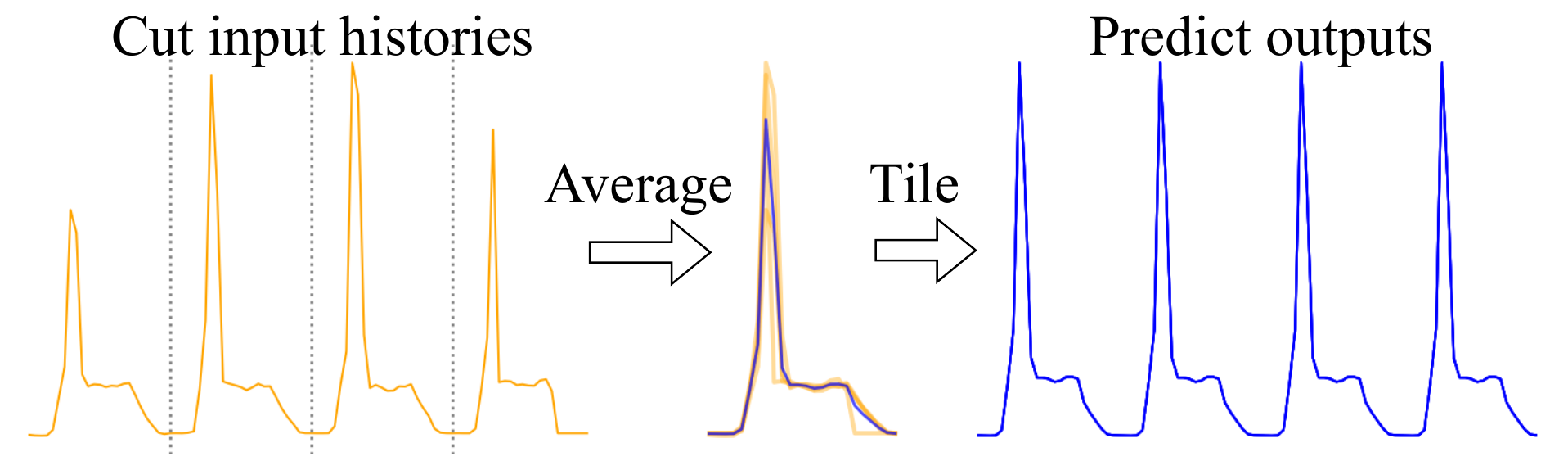}
\caption{The computational steps of \textsf{Average Tile}. First, given the dataset, we determine a single meaningful periodicity $P$. Then, given input histories with length $I$ where $I = r P, r \in \mathbb{Z}^+$, we cut the input histories into $r$ non-overlapping subseries. We then overlap the $r$ subseries and average them series-wise. Finally, we tile the averaged subseries until the appropriate output horizon.}
\label{fig:average_tile}
\end{minipage}

\vspace{-14pt}

\end{figure}

\vspace{-10pt}
\section{The \textsf{AverageTile} Baseline Model} \label{sec:baseline}
\vspace{-10pt}
To design a strong yet simple baseline model, we make a very simple but important observation: there are strong periodicities in most time series datasets.
In Fig.~\ref{fig:period_obs}, we plot part of an univariate series in both time and frequency domain from four common benchmark datasets.
Out of the four datasets, only \texttt{Exchange rate} has no meaningful periodicity.
% \texttt{ETTm2} (Electricity Transformer Temperature) has a periodicity of $96$ because the sampling interval is $15$ minutes and measurements are affected by ambient temperature and daily electricity usage.
% Similarly, hourly-sampled \texttt{Traffic} dataset has daily and weekly cycles and \texttt{Weather} dataset has a periodicity of $144$ because the sampling interval is $10$ minutes with a daily cycle.
In fact, \texttt{Exchange rate} is the only one without clear periodicity out of the seven datasets we benchmark.
Thus, we assume that a time series dataset has at least one (often dominant) periodicity of $P$ time steps that can be exploited for the purpose of long-term forecasting.
As for non-periodic time series, we assume $P = 1$.

Based on this observation, we propose a baseline model, \textsf{AverageTile}, that has zero trainable parameters.
In \textsf{AverageTile}, we need to first determine $P$.
Then, given the input histories $\{\mathbf{x}_{t-I}, \dots, \mathbf{x}_{t-1}\}$ with length $I = r P, r \in \mathbb{Z}^+$, the  future prediction output for horizon $o, o \in [0, \dots, O-1]$ is
\begin{align}
    \mathbf{\hat{x}}_{t+o} = \frac{\sum\limits_{i=1}^{r} \mathbf{x}_{t + (o \text{ mod } P) - i P}}{r}.
\end{align}
In other words, \textsf{AverageTile} cuts the input histories into $r$ non-overlapping subseries with length $P$, then \emph{averages} them series-wise, and finally \emph{tiles} the averaged subseries temporally to the appropriate output horizon.
Notice that there is no trainable parameter in the process and there are no interactions between series.
Conceptually, \textsf{AverageTile} averages the cycles observed in the input and predicts the same averaged cycle for any future time step.
The process is illustrated in Fig.~\ref{fig:average_tile}.

The experimental results of \textsf{AverageTile} compared to \textsf{Autoformer} under two settings are shown in Tab.~\ref{tab:same_results} and Tab.~\ref{tab:best_results}.
Clearly, the baseline model \textsf{AverageTile} has similar performances compared to \textsf{Autoformer}.
Taking what we derived in Sec.~\ref{sec:erracc}, we can see that by focusing entirely on the periodicities, we assume the DGP is more like $x_t = f(t) + e_t$ instead of $x_t = f(x_{t-1}, x_{t-2}, \dots) + e_t$, thus resulting in less error accumulation.
This is akin to typical cycle-seasonal-residual time series decomposition, such as X11, SEATS~\cite{dagum}, or STL~\cite{stl}, where the residuals are close to stationary.
Here, \textsf{AverageTile} models the seasonal part and thus the residual part is closer to stationary.

\vspace{-10pt}
\section{The \textsf{FreDo} Model} \label{sec:fredo}
\vspace{-10pt}

\begin{figure}[t!]
\centering

\includegraphics[width=0.80\columnwidth]{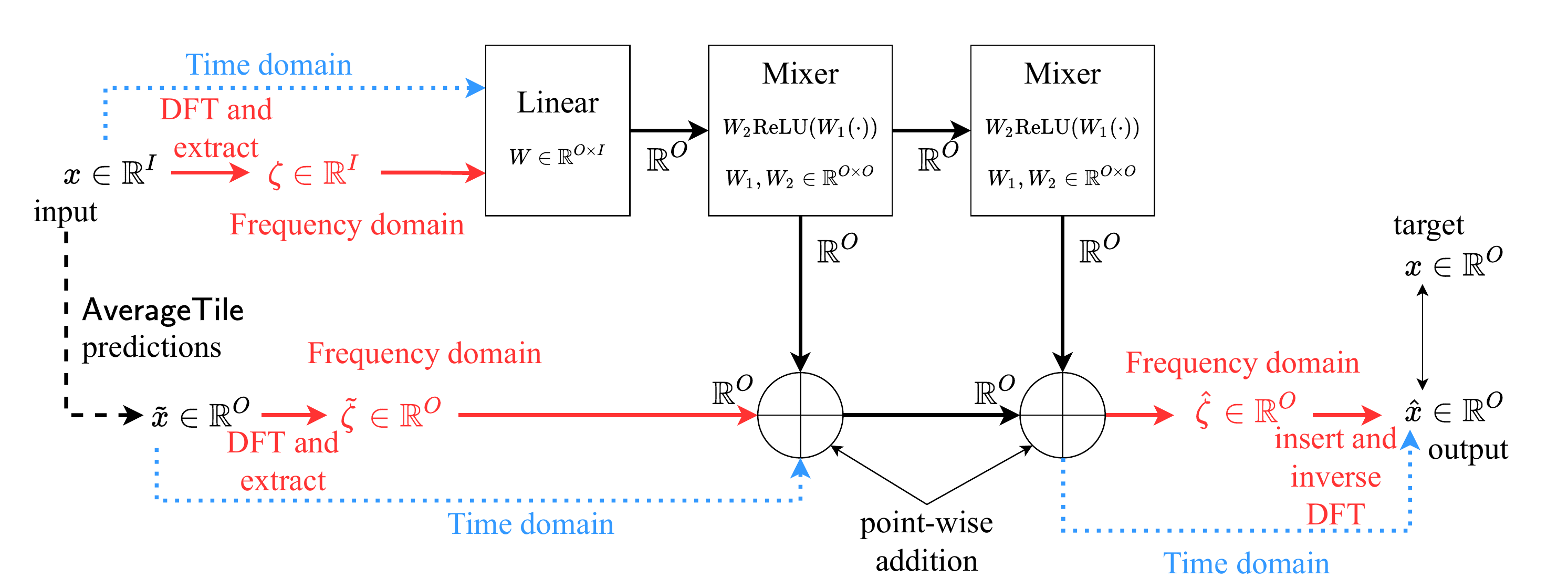}
\caption{Model architecture and data flow for both \textsf{FreDo} and \textsf{TimeDo}. The red and solid lines denote the data flow when the model is trained in frequency domain as in the case of \textsf{FreDo}, and the blue and dotted lines denote the time-domain version of data flow when \textsf{TimeDo} is used.
Conceptually speaking, the design of the model allows it to iteratively refine the baseline prediction from \textsf{AverageTile} by adding the outputs of Mixer modules and thus achieve better performance.}
\label{fig:fredo}

\vspace{-12pt}

\end{figure}

Based upon the strong performance of \textsf{AverageTile}, we propose \textsf{FreDo}, a frequency domain-based model for long-term forecasting.
The idea of using frequency domain is that periodicities can be more easily captured in frequency domain, especially in long-term forecasting where there will be multiple cycles.
The model architecture is shown in Fig.~\ref{fig:fredo}.
\textsf{FreDo} is designed for univariate time series.
For multivariate time series forecasting, the same model is trained and tested on all series jointly.
Thus, for simplicity, we assume the input is univariate with length $I$.
First, given the input $x \in \mathbb{R}^I$, we use Discrete Fourier Transform (DFT) to transform into frequency domain: $\zeta^{\text{raw}} \in \mathbb{C}^I = \text{DFT}(x)$.
The corresponding frequency bin centers are 
\vspace{-6pt}
\begin{align*}
    &I^{-1} [-\frac{I}{2}+1, \dots, -1, 0, 1, \dots, \frac{I}{2}], \text{if } I \text{ is even}, \\
    &I^{-1} [-\frac{I-1}{2}, \dots, -1, 0, 1, \dots, \frac{I-1}{2}], \text{if } I \text{ is odd}. \\[-17pt]
\end{align*}
Because $x$ is real-valued, the negative frequency components are just complex conjugates of the positive frequency components; thus, we only use positive frequency components.
Furthermore, the imaginary part of the frequency component is always zero for frequency $= 0$ and $= \frac{I}{2}$ if $I$ is even; and for frequency $= 0$ if $I$ is odd.
Eventually, we extract all non-trivial numbers from $\zeta^{\text{raw}}$ and concatenate them into a real-valued vector:
\begin{align}
    \zeta = [\Re(\zeta^{\text{raw}}_0), \dots, \Re(\zeta^{\text{raw}}_{\frac{I}{2}}), \Im(\zeta^{\text{raw}}_1), \dots, \Im(\zeta^{\text{raw}}_{\frac{I}{2}-1})] \in \mathbb{R}^I, \text{if } I \text{ is even}, \label{eq:evendft} \\
    \zeta = [\Re(\zeta^{\text{raw}}_0), \dots, \Re(\zeta^{\text{raw}}_{\frac{I-1}{2}}), \Im(\zeta^{\text{raw}}_1), \dots, \Im(\zeta^{\text{raw}}_{\frac{I-1}{2}})] \in \mathbb{R}^I, \text{if } I \text{ is odd}, \label{eq:odddft}
\end{align}
where $\Re$ and $\Im$ denote the real and imaginary part of a complex number.
Notice that in Eq.~(\ref{eq:evendft}) and Eq.~(\ref{eq:odddft}), the $\zeta$ vector still has the same length as the input $x$.
We refer these two operations as ``DFT and extract.''
As for inverse DFT, we can insert the entries in $\zeta$ into the corresponding positions in $\zeta^{\text{raw}}$ and thus obtain the inverse DFT.
We refer to the inverse operations as ``insert and inverse DFT.''

After taking the ``DFT and extract'' operations, $\zeta$ is fed into a linear layer that projects the dimension from $I$ to $O$.
Then, it is fed into several layers of Mixer modules~\cite{mixer} sequentially, where each Mixer module is two linear layers with $O$ hidden dimensions and a ReLU activation function in between.
Finally, we apply the ``insert and inverse DFT'' so the vector is transformed back to time domain where the loss and gradients are calculated.
In summary, the model is trained in the frequency domain but with real values so we do not have to deal with complex-valued training.
In fact, our experimental results show that training in real domain yields better results in most cases as shown in Appendix~B.

We have tried to directly learn the mapping from the input $\{x_{t-I}, \dots, x_{t-1}\}$ to the output $\{x_t, \dots, x_{t+O-1}\}$, but the results are quite poor.
Instead, we decide to build on top of the already good baseline \textsf{AverageTile}.
Thus, we take in the prediction made by \textsf{AverageTile}, apply ``DFT and extract,'' and refine it in frequency domain by adding outputs of the Mixer modules, as shown in Fig.~\ref{fig:fredo}.
The refined prediction is then the output of the model, which also goes through the ``insert and inverse DFT'' phase to obtain the final time domain output.
Essentially, we add learnable parameters to the baseline \textsf{AverageTile} model to further enhance its performance.

Compared to state-of-the-art methods such as \textsf{Informer}~\cite{informer} and  \textsf{Autoformer}~\cite{autoformer} which are highly sophisticated, \textsf{FreDo} is not only much simpler and smaller, but also outperforms by a huge margin.

\vspace{-10pt}
\section{Experimental Results} \label{sec:exp}
\vspace{-10pt}
We extensively perform several experiments on seven public datasets to demonstrate the superiority of our models.
First, we show that the baseline \textsf{AverageTile} model achieves comparable performances against \textsf{Autoformer}.
Next, we compare \textsf{FreDo} against \textsf{Autoformer} under two settings, and see that it significantly outperforms.
Finally, we show that learning in the frequency domain is indeed better by comparing the univariate performance of models in both frequency and time domain.

\setlength\tabcolsep{2.1pt}
% \definecolor{Gray}{gray}{.9}

\begin{table*}[t]
\small
\centering
  \caption{Periodicites of datasets and the chosen $P$ in two settings. In the ``\textsf{Autoformer} setting,'' the input length $I$ is 36 for \texttt{ILI} and 96 for other datasets. Under this constraint, we cannot choose $P$ larger than $I$, thus we use $P = 1$ in \texttt{Weather}, \texttt{Solar}, and \texttt{ILI}, and $P = 24$ in \texttt{Electricity} and \texttt{Traffic}. In the case of ``one-cycle setting,'' we choose $I=P$ so one cycle of data points is observed.}
  \begin{tabular}{l|*{7}{|c}}
  \hline
  Dataset & \texttt{ETTm2} & \texttt{Electricity} & \texttt{Exchange} & \texttt{Traffic} & \texttt{Weather} & \texttt{Solar} & \texttt{ILI} \\
  \hline
  \hline
  Sample interval & 15 min. & 1 hour & 1 day & 1 hour & 10 min. & 10 min. & 1 week \\
  \hline
  Periodicities & daily & daily, weekly & none & daily, weekly & daily & daily & yearly \\
  \hline
  Periodicities in $P$ & 96 & 24, 168 & none & 24, 168 & 144 & 144 & 52 \\
  \hline
  $P$ in ``\textsf{Autoformer} setting'' & 96 & 24 & 1 & 24 & 1 & 1 & 1 \\
  \hline
  $P$ in ``one-cycle setting'' & 96 & 168 & 1 & 168 & 144 & 144 & 52
  \end{tabular}
  \label{tab:settings}
  \vspace{-15pt}
\end{table*}

\begin{figure}[b!]
\centering

\vspace{-14pt}
\includegraphics[width=0.8\columnwidth]{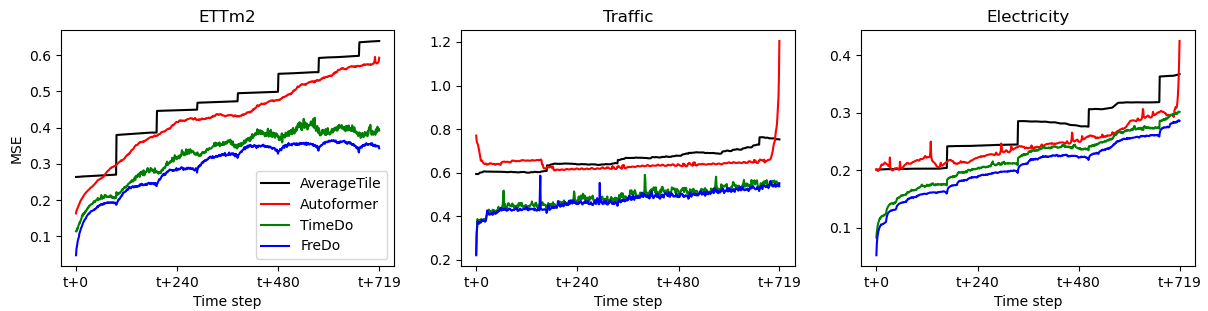}
\caption{The error curves of \textsf{AverageTile}, \textsf{Autoformer}, \textsf{TimeDo}, and \textsf{FreDo} under three datasets (\texttt{ETTm2}, \texttt{Traffic}, \texttt{Electricity}) when output length is 720 with the ``one-cycle setting.'' The error curve measures the errors for the next 720 time steps $\{\hat{x}_t, \dots, \hat{x}_{t+719}\}$ averaged over all possible $t$.}
\label{fig:errcurve}

\end{figure}

\vspace{-10pt}
\subsection{Datasets} \label{sec:datasets}
\vspace{-10pt}
The descriptions of the seven datasets are as follows:
(1) \texttt{ETTm2}~\cite{informer} is collected from electricity transformer temperatures every 15 minutes from July 2016 to July 2018;
(2) \texttt{Electricity}~\cite{LSTNet} records hourly electricity consumption from 2012 to 2014;
(3) \texttt{Exchange}~\cite{LSTNet} includes the daily exchange rate of eight currencies from 1990 to 2016;
(4) \texttt{Traffic} represents hourly road occupancy rates on the San Francisco Bay Area freeways from 2015 to 2016;
(5) \texttt{Weather}\footnote{ https://www.bgc-jena.mpg.de/wetter} records 21 meteorological indicators every 10 minutes from a weather station in Germany;
(6) \texttt{Solar} measures solar power production in 2006 from photovoltaic power plants in Alabama; and
(7) \texttt{ILI}\footnote{
https://gis.cdc.gov/grasp/fluview/fluportaldashboard.html} counts the weekly ratio of recorded influenza-like illness (ILI) patients among all patients from 2002 to 2021.

We follow the setup in \texttt{Autoformer}~\cite{autoformer} and split all datasets into training/validation/testing chronologically by 60\%/20\%/20\% for the \texttt{ETT} and by 70\%/10\%/20\% for other datasets. We also normalize the data according to the mean and variance of the training set for each series independently.
Lastly, we calculate the mean-squared error (MSE) and mean-absolute error (MAE) on the test set to compare different models.

\setlength\tabcolsep{2.1pt}
\definecolor{Gray}{gray}{.9}

\begin{table*}[t]
\small
\centering
  \caption{Results under the ``\textsf{Autoformer} setting.'' The input lengths are 36 for \texttt{ILI} and 96 for other datasets. Four output lengths are tested for each dataset. They are {24, 36, 48, 60} for \texttt{ILI} dataset and {96, 192, 336, 720} for others. Best performance among all models is in boldface and the second best is underlined. The results of \textsf{Autoformer} are referenced directly from their paper~\cite{autocorrelation} except \texttt{Solar}, where we run their released code 5 times and report the average. Under the \textsf{AverageTile} column, numbers with asterisk indicate that the baseline model outperforms the \textsf{Autoformer}.}
  \begin{tabular}{rc||*{3}{cc|}cc}
    \multicolumn{10}{c}{} \\ [-1.9ex]
    \hline
    \multicolumn{10}{c}{} \\ [-1.9ex]
    \multirow{2}{*}{\scriptsize \rotatebox{90}{Dataset}} & \scriptsize Model & \multicolumn{2}{c|}{\textsf{Autoformer}} & \multicolumn{2}{c|}{\textsf{AverageTile}} & \multicolumn{2}{c|}{\textsf{TimeDo}} & \multicolumn{2}{c}{\textsf{FreDo}} \\
    
    \multicolumn{9}{c}{} \\ [-2.45ex]
    & \scriptsize Metric & \scriptsize MSE & \scriptsize MAE & \scriptsize MSE & \scriptsize MAE & {\scriptsize MSE $\pm$ std} & {\scriptsize MAE $\pm$ std} & {\scriptsize MSE $\pm$ std} & {\scriptsize MAE $\pm$ std} \\
    
    \multicolumn{10}{c}{} \\ [-1.9ex]
    \hline
    \multicolumn{10}{c}{} \\ [-1.9ex]

    \multirow{4}{*}{\rotatebox{90}{\texttt{ETTm2}}} & 96 & .255 & .339 & .263 & .301* & \underline{.184 $\pm$ \scriptsize .0018} & \underline{.281 $\pm$ \scriptsize .0020} & \textbf{.169} $\pm$ \scriptsize \textbf{.0011} & \textbf{.268} $\pm$ \scriptsize \textbf{.0007} \\
    & 192 & .281 & .340 & .321 & .337* & \underline{.216 $\pm$ \scriptsize .0023} & \underline{.309 $\pm$ \scriptsize .0023} & \textbf{.200} $\pm$ \scriptsize \textbf{.0006} & \textbf{.295} $\pm$ \scriptsize \textbf{.0004} \\
    & 336 & .339 & .372 & .376 & .370* & \underline{.257 $\pm$ \scriptsize .0018} & \underline{.340 $\pm$ \scriptsize .0013} & \textbf{.237} $\pm$ \scriptsize \textbf{.0006} & \textbf{.323} $\pm$ \scriptsize \textbf{.0003} \\
    & 720 & .422 & .419 & .471 & .422 & \underline{.321 $\pm$ \scriptsize .0031} & \underline{.384 $\pm$ \scriptsize .0039} & \textbf{.297} $\pm$ \scriptsize \textbf{.0032} & \textbf{.366} $\pm$ \scriptsize \textbf{.0011} \\
    \multicolumn{10}{c}{} \\ [-1.8ex]
    \multirow{4}{*}{\rotatebox{90}{\scriptsize \texttt{Electricity}}} & 96 & .201 & .317 & .229 & .290* & \underline{.179 $\pm$ \scriptsize .0006} & \underline{.264 $\pm$ \scriptsize .0004} & \textbf{.178} $\pm$ \scriptsize \textbf{.0004} & \textbf{.262} $\pm$ \scriptsize \textbf{.0001} \\
    & 192 & .222 & .334 & .228 & .294* & \underline{.180 $\pm$ \scriptsize .0002} & \underline{.267 $\pm$ \scriptsize .0001} & \textbf{.179} $\pm$ \scriptsize \textbf{.0002} & \textbf{.266} $\pm$ \scriptsize \textbf{.0001} \\
    & 336 & .231 & .338 & .240 & .307* & \underline{.191 $\pm$ \scriptsize .0002} & \underline{.281 $\pm$ \scriptsize .0003} & \textbf{.190} $\pm$ \scriptsize \textbf{.0002} & \textbf{.280} $\pm$ \scriptsize \textbf{.0000} \\
    & 720 & .254 & .361 & .281 & .338* & \underline{.223 $\pm$ \scriptsize .0006} & \underline{.310 $\pm$ \scriptsize .0002} & \textbf{.221} $\pm$ \scriptsize \textbf{.0006} & \textbf{.309} $\pm$ \scriptsize \textbf{.0004} \\
    \multicolumn{10}{c}{} \\ [-1.8ex]
    \multirow{4}{*}{\rotatebox{90}{\texttt{ Exchange}}} & 96 & .197 & .323 & .139* & .269* & \underline{.096 $\pm$ \scriptsize .0057} & \underline{.228 $\pm$ \scriptsize .0070} & \textbf{.080} $\pm$ \scriptsize \textbf{.0008} & \textbf{.200} $\pm$ \scriptsize \textbf{.0012} \\
    & 192 & .300 & .369 & .235* & .352* & \underline{.191 $\pm$ \scriptsize .0046} & \underline{.324 $\pm$ \scriptsize .0042} & \textbf{.161} $\pm$ \scriptsize \textbf{.0041} & \textbf{.292} $\pm$ \scriptsize \textbf{.0037} \\
    & 336 & .509 & .524 & .383* & \underline{.454*} & \underline{.378 $\pm$ \scriptsize .0325} & .459 $\pm$ \scriptsize .0180 & \textbf{.292} $\pm$ \scriptsize \textbf{.0196} & \textbf{.401} $\pm$ \scriptsize \textbf{.0131} \\
    & 720 & 1.447 & .941 & .931* & .735* & \underline{.793 $\pm$ \scriptsize .1157} & \underline{.695 $\pm$ \scriptsize .0435} & \textbf{.534} $\pm$ \scriptsize \textbf{.0345} & \textbf{.571} $\pm$ \scriptsize \textbf{.0109} \\
    \multicolumn{10}{c}{} \\ [-1.8ex]
    \multirow{4}{*}{\rotatebox{90}{\texttt{ Traffic}}} & 96 & .613 & .388 & .831 & .434 & \underline{.526 $\pm$ \scriptsize .0014} & \underline{.333 $\pm$ \scriptsize .0005} & \textbf{.512} $\pm$ \scriptsize \textbf{.0025} & \textbf{.332} $\pm$ \scriptsize \textbf{.0007} \\
    & 192 & .616 & .382 & .762 & .413 & \underline{.514 $\pm$ \scriptsize .0013} & \textbf{.315} $\pm$ \scriptsize \textbf{.0006} & \textbf{.492} $\pm$ \scriptsize \textbf{.0007} & \underline{.315 $\pm$ \scriptsize .0005} \\
    & 336 & .622 & .337 & .768 & .414 & \underline{.524 $\pm$ \scriptsize .0006} & \textbf{.312} $\pm$ \scriptsize \textbf{.0002} & \textbf{.500} $\pm$ \scriptsize \textbf{.0005} & \underline{.312 $\pm$ \scriptsize .0004} \\
    & 720 & .660 & .408 & .806 & .429 & \underline{.564 $\pm$ \scriptsize .0029} & \textbf{.323} $\pm$ \scriptsize \textbf{.0004} & \textbf{.538} $\pm$ \scriptsize \textbf{.0066} & \underline{.323 $\pm$ \scriptsize .0007} \\
    \multicolumn{10}{c}{} \\ [-1.8ex]
    \multirow{4}{*}{\rotatebox{90}{\texttt{Weather}}} & 96 & .266 & .336 & .215* & .271* & \underline{.183 $\pm$ \scriptsize .0041} & \underline{.238 $\pm$ \scriptsize .0008} & \textbf{.171} $\pm$ \scriptsize \textbf{.0008} & \textbf{.227} $\pm$ \scriptsize \textbf{.0014} \\
    & 192 & .307 & .367 & .264* & .305* & \underline{.224 $\pm$ \scriptsize .0014} & \underline{.273 $\pm$ \scriptsize .0006} & \textbf{.211} $\pm$ \scriptsize \textbf{.0008} & \textbf{.264} $\pm$ \scriptsize \textbf{.0021} \\
    & 336 & .359 & .395 & .312* & .336* & \underline{.269 $\pm$ \scriptsize .0009} & \underline{.308 $\pm$ \scriptsize .0010} & \textbf{.260} $\pm$ \scriptsize \textbf{.0010} & \textbf{.301} $\pm$ \scriptsize \textbf{.0026} \\
    & 720 & .419 & .428 & .381* & .377* & \underline{.336 $\pm$ \scriptsize .0033} & \underline{.361 $\pm$ \scriptsize .0038} & \textbf{.331} $\pm$ \scriptsize \textbf{.0013} & \textbf{.347} $\pm$ \scriptsize \textbf{.0020} \\
    \multicolumn{10}{c}{} \\ [-1.8ex]
    \multirow{4}{*}{\rotatebox{90}{\texttt{Solar}}} & 96 & .408 & .421 & .911 & .734 & \textbf{.215} $\pm$ \scriptsize \textbf{.0047} & \textbf{.270} $\pm$ \scriptsize \textbf{.0029} & \underline{.216 $\pm$ \scriptsize .0034} & \underline{.276 $\pm$ \scriptsize .0030} \\
    & 192 & .763 & .649 & .925 & .741 & \textbf{.228} $\pm$ \scriptsize \textbf{.0034} & \textbf{.281} $\pm$ \scriptsize \textbf{.0022} & \underline{.239 $\pm$ \scriptsize .0063} & \underline{.288 $\pm$ \scriptsize .0033} \\
    & 336 & 1.102 & 0.914 & .915* & .734* & \textbf{.229} $\pm$ \scriptsize \textbf{.0032} & \textbf{.278} $\pm$ \scriptsize \textbf{.0049} & \underline{.237 $\pm$ \scriptsize .0040} & \underline{.286 $\pm$ \scriptsize .0050} \\
    & 720 & .678 & .603 & .879 & .715 & \textbf{.229} $\pm$ \scriptsize \textbf{.0031} & \textbf{.274} $\pm$ \scriptsize \textbf{.0045} & \underline{.230 $\pm$ \scriptsize .0036} & \underline{.279 $\pm$ \scriptsize .0027} \\
    \multicolumn{10}{c}{} \\ [-1.8ex]
    \multirow{4}{*}{\rotatebox{90}{\texttt{ILI}}} & 24 & 3.483 & \underline{1.287} & 6.643 & 1.953 & \textbf{2.944} $\pm$ \scriptsize \textbf{.0875} & \textbf{1.129} $\pm$ \scriptsize \textbf{.0256} & \underline{3.286 $\pm$ \scriptsize .0592} & 1.307 $\pm$ \scriptsize .0160 \\
    & 36 & \textbf{3.103} & \textbf{1.148} & 6.267 & 1.889 & 3.329 $\pm$ \scriptsize .0682 & \underline{1.210 $\pm$ \scriptsize .0132} & \underline{3.252 $\pm$ \scriptsize .0505} & 1.279 $\pm$ \scriptsize .0148 \\
    & 48 & \textbf{2.669} & \textbf{1.085} & 5.444 & 1.693 & 3.438 $\pm$ \scriptsize .0917 & \underline{1.231 $\pm$ \scriptsize .0179} & \underline{3.298 $\pm$ \scriptsize .0209} & 1.275 $\pm$ \scriptsize .0054 \\
    & 60 & \textbf{2.770} & \textbf{1.125} & 4.798 & 1.593 & 3.645 $\pm$ \scriptsize .0561 & \underline{1.278 $\pm$ \scriptsize .0098} & \underline{4.480 $\pm$ \scriptsize .0110} & 1.303 $\pm$ \scriptsize .0036
  \end{tabular}
  \label{tab:same_results}
  \vspace{-20pt}
\end{table*}

\vspace{-5pt}
\subsection{Settings} \label{sec:settings}
\vspace{-5pt}
To exhibit the strong performance of \textsf{AverageTile} and the even stronger performance of \textsf{FreDo}, we compare our models to the state-of-the-art \textsf{Autoformer} model~\cite{autoformer}.
We do not compare against other models such as \textsf{Informer}~\cite{informer}, \textsf{Reformer}~\cite{reformer}, or \textsf{LogTrans}~\cite{logtrans} because \textsf{Autoformer} has shown to greatly outperform those models.
Additionally, to demonstrate the advantage of frequency domain learning, we also show the results of  \textsf{TimeDo}, the time-domain version of \textsf{FreDo}.
That is, without applying the ``DFT and extract,'' we directly take the input along with the baseline prediction and feed it into the model.
The output is also used directly without the ``insert and inverse DFT'' step.
The number of parameters of \textsf{FreDo} and \textsf{TimeDo} are thus exactly the same.

To compare against \textsf{Autoformer} fairly but to also show how low an error can be achieved using our models, we devise two settings.
In both settings, all setups except the input length $I$ follow exactly the \textsf{Autoformer}~\cite{autocorrelation} paper and have been double checked with their released code.
% The only difference between the two settings is the input length $I$ and subsequently the periodicity $P$ for each dataset.
The first setting is the same as in the \textsf{Autoformer} paper~\cite{autoformer}; that is, $I = 36$ for \texttt{ILI} and $I = 96$ for other datasets, in order to compare fairly against the results reported in their paper.
However, since the $P$ for some datasets are longer, we further experiment with one cycle of input (i.e., $I = P$) to showcase the lowest error rate our models can achieve.
We call the first setting the ``\textsf{Autoformer} setting,'' and name the second the ``one-cycle setting.''
The summary of both settings can be found in Tab.~\ref{tab:settings}.
In addition, detailed hyperparameters of our models can be found in Appendix~A.

\setlength\tabcolsep{1.7pt}
\definecolor{Gray}{gray}{.9}

\begin{table*}[t]
\small
\centering
  \caption{Results under the ``one-cycle setting.''  The input lengths are set to the largest $P$ as shown in Tab.~\ref{tab:settings}, so exactly one-cycle of data is given as input. Best performance among all models is in boldface and the second best is underlined. The results of \textsf{Autoformer} are collected by running their released code by 5 times. Under the \textsf{AverageTile} column, numbers with asterisk indicate that the baseline model outperforms the \textsf{Autoformer}. Lower MSE or MAE is better.}
  \begin{tabular}{rc||*{3}{cc|}cc}
    \multicolumn{10}{c}{} \\ [-1.9ex]
    \hline
    \multicolumn{10}{c}{} \\ [-1.9ex]
    \multirow{2}{*}{\scriptsize \rotatebox{90}{Dataset}} & \scriptsize Model & \multicolumn{2}{c|}{\textsf{Autoformer}} & \multicolumn{2}{c|}{\textsf{AverageTile}} & \multicolumn{2}{c|}{\textsf{TimeDo}} & \multicolumn{2}{c}{\textsf{FreDo}} \\
    
    \multicolumn{9}{c}{} \\ [-2.45ex]
    & \scriptsize Metric & \scriptsize MSE $\pm$ std & \scriptsize MAE $\pm$ std & \scriptsize MSE & \scriptsize MAE & {\scriptsize MSE $\pm$ std} & {\scriptsize MAE $\pm$ std} & {\scriptsize MSE $\pm$ std} & {\scriptsize MAE $\pm$ std} \\
    
    \multicolumn{10}{c}{} \\ [-2.2ex]
    \hline
    \multicolumn{10}{c}{} \\ [-2.2ex]

    \multirow{4}{*}{\rotatebox{90}{\texttt{ETTm2}}} & 96 & .255 $\pm$ \scriptsize .0187 & .327 $\pm$ \scriptsize .0101 & .263 & .300* & \underline{.199 $\pm$ \scriptsize .0192} & \underline{.288 $\pm$ \scriptsize .0086} & \textbf{.178} $\pm$ \scriptsize \textbf{.0117} & \textbf{.270} $\pm$ \scriptsize \textbf{.0023} \\
    & 192 & .293 $\pm$ \scriptsize .0083 & .346 $\pm$ \scriptsize .0062 & .321 & .337* & \underline{.244 $\pm$ \scriptsize .0351} & \underline{.322 $\pm$ \scriptsize .0167} & \textbf{.223} $\pm$ \scriptsize \textbf{.0281} & \textbf{.304} $\pm$ \scriptsize \textbf{.0105} \\
    & 336 & .346 $\pm$ \scriptsize .0087 & .378 $\pm$ \scriptsize .0032 & .376 & .370* & \underline{.293 $\pm$ \scriptsize .0435} & \underline{.356 $\pm$ \scriptsize .0190} & \textbf{.270} $\pm$ \scriptsize \textbf{.0404} & \textbf{.337} $\pm$ \scriptsize \textbf{.0172} \\
    & 720 & .447 $\pm$ \scriptsize .0270 & .433 $\pm$ \scriptsize .0138 & .471 & .421* & \underline{.385 $\pm$ \scriptsize .0842} & \underline{.417 $\pm$ \scriptsize .0436} & \textbf{.345} $\pm$ \scriptsize \textbf{.0558} & \textbf{.390} $\pm$ \scriptsize \textbf{.0295} \\
    \multicolumn{10}{c}{} \\ [-1.8ex]
    \multirow{4}{*}{\rotatebox{90}{\scriptsize \texttt{Electricity}}} & 96 & .196 $\pm$ \scriptsize .0023 & .311 $\pm$ \scriptsize .0027 & .212 & .279* & \underline{.160 $\pm$ \scriptsize .0002} & \underline{.250 $\pm$ \scriptsize .0002} & \textbf{.148} $\pm$ \scriptsize \textbf{.0004} & \textbf{.240} $\pm$ \scriptsize \textbf{.0001} \\
    & 192 & .211 $\pm$ \scriptsize .0064 & .324 $\pm$ \scriptsize .0053 & .216 & .283* & \underline{.173 $\pm$ \scriptsize .0003} & \underline{.261 $\pm$ \scriptsize .0003} & \textbf{.160} $\pm$ \scriptsize \textbf{.0005} & \textbf{.251} $\pm$ \scriptsize \textbf{.0002} \\
    & 336 & .236 $\pm$ \scriptsize .0257 & .342 $\pm$ \scriptsize .0142 & .228* & .295* & \underline{.187 $\pm$ \scriptsize .0002} & \underline{.277 $\pm$ \scriptsize .0002} & \textbf{.173} $\pm$ \scriptsize \textbf{.0004} & \textbf{.266} $\pm$ \scriptsize \textbf{.0004} \\
    & 720 & .267 $\pm$ \scriptsize .0237 & .365 $\pm$ \scriptsize .0128 & .267 & .325 & \underline{.219 $\pm$ \scriptsize .0013} & \underline{.307 $\pm$ \scriptsize .0005} & \textbf{.206} $\pm$ \scriptsize \textbf{.0006} & \textbf{.295} $\pm$ \scriptsize \textbf{.0006} \\
    \multicolumn{10}{c}{} \\ [-1.8ex]
    \multirow{4}{*}{\rotatebox{90}{\texttt{ Exchange}}} & 96 & .085 $\pm$ \scriptsize .0010 & .203 $\pm$ \scriptsize .0015 & .081* & \underline{.196*} & \underline{.080 $\pm$ \scriptsize .0006} & .198 $\pm$ \scriptsize .0012 & \textbf{.078} $\pm$ \scriptsize \textbf{.0004} & \textbf{.195} $\pm$ \scriptsize \textbf{.0005} \\
    & 192 & .185 $\pm$ \scriptsize .0043 & .309 $\pm$ \scriptsize .0038 & .167* & \underline{.289*} & \underline{.164 $\pm$ \scriptsize .0031} & .295 $\pm$ \scriptsize .0026 & \textbf{.154} $\pm$ \scriptsize \textbf{.0022} & \textbf{.285} $\pm$ \scriptsize \textbf{.0017} \\
    & 336 & .341 $\pm$ \scriptsize .0102 & .425 $\pm$ \scriptsize .0062 & \underline{.306*} & \underline{.398*} & .317 $\pm$ \scriptsize .0128 & .417 $\pm$ \scriptsize .0074 & \textbf{.257} $\pm$ \scriptsize \textbf{.0024} & \textbf{.378} $\pm$ \scriptsize \textbf{.0023} \\
    & 720 & 1.012 $\pm$ \scriptsize .0234 & .775 $\pm$ \scriptsize .0113 & .810* & .676* & \underline{.559 $\pm$ \scriptsize .0426} & \underline{.574 $\pm$ \scriptsize .0196} & \textbf{.490} $\pm$ \scriptsize \textbf{.0024} & \textbf{.541} $\pm$ \scriptsize \textbf{.0032} \\
    \multicolumn{10}{c}{} \\ [-1.8ex]
    \multirow{4}{*}{\rotatebox{90}{\texttt{ Traffic}}} & 96 & .630 $\pm$ \scriptsize .0207 & .392 $\pm$ \scriptsize .0119 & .620* & \textbf{.296}* & \underline{.456 $\pm$ \scriptsize .0010} & .303 $\pm$ \scriptsize .0005 & \textbf{.445} $\pm$ \scriptsize \textbf{.0010} & \underline{.299 $\pm$ \scriptsize .0004} \\
    & 192 & .625 $\pm$ \scriptsize .0246 & .387 $\pm$ \scriptsize .0218 & .624* & \textbf{.297*} & \underline{.457 $\pm$ \scriptsize .0017} & .302 $\pm$ \scriptsize .0003 & \textbf{.446} $\pm$ \scriptsize \textbf{.0010} & \underline{.299 $\pm$ \scriptsize .0006} \\
    & 336 & .619 $\pm$ \scriptsize .0138 & .382 $\pm$ \scriptsize .0118 & .632 & .302* & \underline{.461 $\pm$ \scriptsize .0024} & \underline{.300 $\pm$ \scriptsize .0006} & \textbf{.454} $\pm$ \scriptsize \textbf{.0012} & \textbf{.299} $\pm$ \scriptsize \textbf{.0004} \\
    & 720 & .644 $\pm$ \scriptsize .0108 & .395 $\pm$ \scriptsize .0080 & .660 & .319* & \underline{.494 $\pm$ \scriptsize .0032} & \underline{.311 $\pm$ \scriptsize .0013} & \textbf{.474} $\pm$ \scriptsize \textbf{.0010} & \textbf{.309} $\pm$ \scriptsize \textbf{.0006} \\
    \multicolumn{10}{c}{} \\ [-1.8ex]
    \multirow{4}{*}{\rotatebox{90}{\texttt{Weather}}} & 96 & .248 $\pm$ \scriptsize .0063 & .327 $\pm$ \scriptsize .0076 & .317 & .288* & \underline{.196 $\pm$ \scriptsize .0010} & \underline{.235 $\pm$ \scriptsize .0008} & \textbf{.194} $\pm$ \scriptsize \textbf{.0009} & \textbf{.227} $\pm$ \scriptsize \textbf{.0006} \\
    & 192 & .339 $\pm$ \scriptsize .0121 & .396 $\pm$ \scriptsize .0104 & .343 & .305* & \underline{.234 $\pm$ \scriptsize .0004} & \underline{.270 $\pm$ \scriptsize .0009} & \textbf{.230} $\pm$ \scriptsize \textbf{.0011} & \textbf{.262} $\pm$ \scriptsize \textbf{.0008} \\
    & 336 & .354 $\pm$ \scriptsize .0108 & .393 $\pm$ \scriptsize .0094 & .383 & .331* & \underline{.282 $\pm$ \scriptsize .0015} & \underline{.308 $\pm$ \scriptsize .0012} & \textbf{.276} $\pm$ \scriptsize \textbf{.0008} & \textbf{.298} $\pm$ \scriptsize \textbf{.0006} \\
    & 720 & .468 $\pm$ \scriptsize .0178 & .469 $\pm$ \scriptsize .0143 & .443* & .370* & \underline{.353 $\pm$ \scriptsize .0023} & \underline{.360 $\pm$ \scriptsize .0033} & \textbf{.346} $\pm$ \scriptsize \textbf{.0008} & \textbf{.348} $\pm$ \scriptsize \textbf{.0016} \\
    \multicolumn{10}{c}{} \\ [-1.8ex]
    \multirow{4}{*}{\rotatebox{90}{\texttt{Solar}}} & 96 & .466 $\pm$ \scriptsize .0487 & .467 $\pm$ \scriptsize .0201 & .290* & \textbf{.224*} & \textbf{.175} $\pm$ \scriptsize \textbf{.0033} & \underline{.234 $\pm$ \scriptsize .0033} & \underline{.176 $\pm$ \scriptsize .0008} & .234 $\pm$ \scriptsize .0015 \\
    & 192 & .761 $\pm$ \scriptsize .1948 & .618 $\pm$ \scriptsize .0895 & .327* & \textbf{.243*} & \textbf{.192} $\pm$ \scriptsize \textbf{.0027} & .250 $\pm$ \scriptsize .0023 & \underline{.193 $\pm$ \scriptsize .0012} & \underline{.248 $\pm$ \scriptsize .0008} \\
    & 336 & .820 $\pm$ \scriptsize .0940 & .690 $\pm$ \scriptsize .0369 & .372* & .266* & \textbf{.199} $\pm$ \scriptsize \textbf{.0018} & \underline{.257 $\pm$ \scriptsize .0017} & \underline{.202 $\pm$ \scriptsize .0018} & \textbf{.255} $\pm$ \scriptsize \textbf{.0019} \\
    & 720 & .834 $\pm$ \scriptsize .1498 & .653 $\pm$ \scriptsize .0576 & .376* & .270* & \textbf{.206} $\pm$ \scriptsize \textbf{.0029} & \underline{.261 $\pm$ \scriptsize .0009} & \underline{.207 $\pm$ \scriptsize .0016} & \textbf{.260} $\pm$ \scriptsize \textbf{.0009} \\
    \multicolumn{10}{c}{} \\ [-1.8ex]
    \multirow{4}{*}{\rotatebox{90}{\texttt{ILI}}} & 24 & 2.964 $\pm$ \scriptsize .2119 & 1.152 $\pm$ \scriptsize .0492 & 2.551* & \underline{1.000*} & \underline{2.468 $\pm$ \scriptsize .0379} & 1.049 $\pm$ \scriptsize .0110 & \textbf{2.293} $\pm$ \scriptsize \textbf{.0291} & \textbf{.988} $\pm$ \scriptsize \textbf{.0070} \\
    & 36 & 2.787 $\pm$ \scriptsize .1849 & 1.082 $\pm$ \scriptsize .0450 & \underline{2.261*} & \textbf{.956*} & 2.610 $\pm$ \scriptsize .0607 & 1.014 $\pm$ \scriptsize .0177 & \textbf{2.239} $\pm$ \scriptsize \textbf{.0277} & \underline{.985 $\pm$ \scriptsize .0079} \\
    & 48 & 2.879 $\pm$ \scriptsize .1986 & 1.112 $\pm$ \scriptsize .0487 & \textbf{2.136*} & \textbf{.939*} & 2.877 $\pm$ \scriptsize .0404 & 1.001 $\pm$ \scriptsize .0178 & \underline{2.304 $\pm$ \scriptsize .0301} & \underline{.974 $\pm$ \scriptsize .0136} \\
    & 60 & 2.779 $\pm$ \scriptsize .1393 & 1.108 $\pm$ \scriptsize .0306 & \textbf{1.911*} & \textbf{.936*} & 3.086 $\pm$ \scriptsize .0596 & \underline{.977 $\pm$ \scriptsize .0112} & \underline{2.316 $\pm$ \scriptsize .0177} & 1.025 $\pm$ \scriptsize .0041
  \end{tabular}
  \label{tab:best_results}
  
  \vspace{-20pt}
  
\end{table*}

\vspace{-10pt}
\subsection{Comparing \textsf{AverageTile} to \textsf{Autoformer}}
\vspace{-5pt}
The results under two settings are shown in Tab.~\ref{tab:same_results} and Tab.~\ref{tab:best_results}.
For both settings, first notice that the baseline model consistently outperforms on the \texttt{Exchange} dataset, which is the only dataset without clear periodicity.
Since $P = 1$ for \texttt{Exchange} dataset, this implies that taking the average of the last $I$ observations as output already outperforms \textsf{Autoformer}.
This further confirms the fact that a sophisticated model can actually underperform a baseline model in long-term forecasting.
Other than the \texttt{Exchange} dataset, \textsf{AverageTile} also outperforms occasionally on other datasets.

To understand why \textsf{AverageTile} performs comparably against \textsf{Autoformer}, we look at the error curves as shown in Fig.~\ref{fig:errcurve}.
The error curves of \textsf{AverageTile} are clear demonstrations of error accumulation in long-term forecasting.
As described in Sec.~\ref{sec:erracc}, a good model should have an increasing error curve.
The observed error curves suggest that \textsf{AverageTile} is a reasonable, good baseline model.
However, the error curves of \textsf{Autoformer} indicate that it has room for improvement, because the left-end of the error curve is inverted and the right-end of the curve explodes, especially on the \texttt{Traffic} and \texttt{Electricity} datasets.
In fact, its errors are higher than \textsf{AverageTile} for the first 100+ time steps.

Furthermore, although \textsf{AverageTile} does not outperform in every case, we emphasize that it is a much simpler model than \textsf{Autoformer} in many aspects.
Specifically, \textsf{AverageTile} has no learnable parameter, no training, and is univariate without any complex, hand-crafted architecture.
In fact, if we search over the hyperparameter $r = \frac{I}{P}$ and choose the best $r$ for each dataset, \textsf{AverageTile} can achieve even better performance.
This extra experiment is detailed in Appendix~C.
% In Appendix~\ref{app:searchbase}, we search over $r = \{1, \dots, 10, \infty \}$, where $\infty$ means that the whole training set is treated as input for any prediction, and show that this improves performance.
It is much more difficult to search hyperparameters for sophisticated models like \textsf{Autoformer} because it is more computationally intensive and the model size increases with longer input length.

\setlength\tabcolsep{4.6pt}
\definecolor{Gray}{gray}{.9}

\begin{table*}[t]
\small
\centering
  \begin{minipage}{.47\linewidth}
      \begin{tabular}{rc||cc|cc}
      
      \multirow{2}{*}{\tiny \rotatebox{90}{Dataset}} & \scriptsize Model & \multicolumn{2}{c|}{\textsf{TimeDo}} & \multicolumn{2}{c}{\textsf{FreDo}} \\
      & \scriptsize Metric & MSE & MAE & MSE & MAE \\
      \hline
      
      \multirow{4}{*}{\rotatebox{90}{\texttt{ETTm2}}} & 96 & 0.249 & 0.320 & $\textbf{0.201}^\dagger$ & $\textbf{0.281}^\dagger$ \\
      & 192 & 0.387 & 0.399 & $\textbf{0.273}^\dagger$ & \textbf{0.332} \\
      & 336 & 0.522 & 0.462 & $\textbf{0.337}^\dagger$ & \textbf{0.368} \\
      & 720 & 1.137 & 0.678 & $\textbf{0.487}^\dagger$ & \textbf{0.458} \\
      \multicolumn{6}{c}{} \\ [-1.8ex]
      \multirow{4}{*}{\rotatebox{90}{\scriptsize \texttt{Electricity}}} & 96 & 0.158 & 0.249 & $\textbf{0.148}^\dagger$ & $\textbf{0.241}^\dagger$ \\
      & 192 & 0.179 & 0.267 & $\textbf{0.166}^\dagger$ & $\textbf{0.257}^\dagger$ \\
      & 336 & 0.204 & 0.287 & $\textbf{0.183}^\dagger$ & $\textbf{0.275}^\dagger$ \\
      & 720 & 0.242 & 0.322 & $\textbf{0.220}^\dagger$ & $\textbf{0.306}^\dagger$ \\
      \multicolumn{6}{c}{} \\ [-1.8ex]
      \multirow{4}{*}{\rotatebox{90}{\texttt{ Exchange}}} & 96 & 0.082 & 0.205 & \textbf{0.082} & \textbf{0.202} \\
      & 192 & 0.169 & 0.304 & \textbf{0.168} & \textbf{0.298} \\
      & 336 & 0.335 & 0.448 & $\textbf{0.282}^\dagger$ & $\textbf{0.403}^\dagger$ \\
      & 720 & 1.390 & 0.857 & $\textbf{0.661}^\dagger$ & $\textbf{0.641}^\dagger$ \\
      \multicolumn{6}{c}{} \\ [-1.8ex]
      \multirow{4}{*}{\rotatebox{90}{\texttt{ Traffic}}} & 96 & 0.523 & 0.318 & $\textbf{0.502}^\dagger$ & $\textbf{0.316}^\dagger$ \\
      & 192 & 0.529 & 0.325 & $\textbf{0.502}^\dagger$ & $\textbf{0.317}^\dagger$ \\
      & 336 & 0.540 & 0.335 & $\textbf{0.526}^\dagger$ & $\textbf{0.328}^\dagger$ \\
      & 720 & 0.561 & 0.356 & $\textbf{0.551}^\dagger$ & $\textbf{0.349}^\dagger$ \end{tabular}
    \end{minipage} \hspace{-10pt}%
    \begin{minipage}{.53\linewidth}
      
      \centering
      \caption{Results of univariately evaluating \textsf{TimeDo} and \textsf{FreDo} on each series. Averaged numbers over all series are reported. Paired t-test is conducted for each pair of models on the same series. Best performance is in boldface and numbers supersribed with $\dagger$ indicates a $p$-value smaller $< 1\%$.%\vspace{2pt}
      }
      \label{tab:single}
      \begin{tabular}{rc||cc|cc}
      
      \multirow{2}{*}{\tiny \rotatebox{90}{Dataset}} & \scriptsize Model & \multicolumn{2}{c|}{\textsf{TimeDo}} & \multicolumn{2}{c}{\textsf{FreDo}} \\
      & \scriptsize Metric & MSE & MAE & MSE & MAE \\
      \hline
      \multirow{4}{*}{\rotatebox{90}{\texttt{Weather}}} & 96 & 0.180 & 0.233 & $\textbf{0.171}^\dagger$ & $\textbf{0.217}^\dagger$ \\
      & 192 & 0.222 & 0.269 & $\textbf{0.210}^\dagger$ & $\textbf{0.255}^\dagger$ \\
      & 336 & 0.273 & 0.308 & $\textbf{0.259}^\dagger$ & $\textbf{0.292}^\dagger$ \\
      & 720 & 0.354 & 0.364 & $\textbf{0.331}^\dagger$ & $\textbf{0.341}^\dagger$ \\
      \multicolumn{6}{c}{} \\ [-1.8ex]
      \multirow{4}{*}{\rotatebox{90}{\texttt{Solar}}} & 96 & 0.193 & 0.256 & $\textbf{0.189}^\dagger$ & $\textbf{0.253}^\dagger$ \\
      & 192 & 0.213 & 0.273 & $\textbf{0.208}^\dagger$ & $\textbf{0.268}^\dagger$ \\
      & 336 & 0.223 & 0.281 & $\textbf{0.220}^\dagger$ & $\textbf{0.275}^\dagger$ \\
      & 720 & 0.228 & 0.285 & $\textbf{0.225}^\dagger$ & $\textbf{0.280}^\dagger$ \\
      \multicolumn{6}{c}{} \\ [-1.8ex]
      \multirow{4}{*}{\rotatebox{90}{\texttt{ILI}}} & 24 & 2.700 & 1.098 & $\textbf{2.225}^\dagger$ & $\textbf{0.920}^\dagger$ \\
      & 36 & 2.815 & 1.127 & $\textbf{2.206}^\dagger$ & $\textbf{0.955}^\dagger$ \\
      & 48 & 3.199 & 1.116 & $\textbf{2.307}^\dagger$ & $\textbf{0.978}^\dagger$ \\
      & 60 & 3.307 & 1.198 & $\textbf{2.335}^\dagger$ & $\textbf{1.008}^\dagger$
      \end{tabular}
  \end{minipage}
  \vspace{-18pt}
\end{table*}

\vspace{-5pt}
\subsection{Performance of \textsf{FreDo}}
\vspace{-5pt}
From both Tab.~\ref{tab:same_results} and Tab.~\ref{tab:best_results}, we can clearly see the superiority of \textsf{FreDo}.
Compared to \textsf{Autoformer}, under both settings, \textsf{FreDo} outperforms by a large margin, with at least 20\% improvement in most cases and up to 50\% improvement in some cases.
Furthermore, the standard deviation is also smaller in most cases.
This is expected because the size of \textsf{Autoformer} is much larger, as tabulated in Appendix~D, so there is higher chance of overfitting and thus higher variance.
Lastly, \textsf{FreDo} uses the same set of parameters for all series in a dataset (which is also the case for \textsf{AverageTile}), so this also reduces overfitting and forces the model to learn more general patterns across all the dataset.
In short, \textsf{FreDo} is a more effective model than \textsf{Autoformer} because it achieves both significantly lower bias and lower variance.
This is again validated by the error curves in Fig.~\ref{fig:errcurve}, where \textsf{FreDo} demonstrates slower rate of error accumulation.

\textsf{TimeDo} also outperforms \textsf{Autoformer} significantly.
Comparing \textsf{FreDo} with \textsf{TimeDo}, we see that \textsf{FreDo} achieves lower errors in most cases.
However, the improvement is statistically insignificant in many cases and even underperforms on the \texttt{Solar} dataset.
Thus, we further perform a detailed experiment to prove that the frequency domain-based model is indeed better.

\vspace{-5pt}
\subsection{Frequency v.s. Time Domain}
\vspace{-5pt}
Recall that both \textsf{TimeDo} and \textsf{FreDo} are univariate models; we are able to run on multivariate datasets because we use the same set of model parameters for each series (each variable or feature).
Thus, the multivariate results in Tab.~\ref{tab:same_results} and Tab.~\ref{tab:best_results} do not provide a complete picture of model capability, because the same univariate model is trying to minimize the errors on all series simultaneously.
Instead, we can train the models, both \textsf{TimeDo} and \textsf{FreDo}, with each univariate series independently, to assess full capability of these models.
For example, there are 321 series (variables or features) in the \texttt{Electricity} dataset; thus, we can train 321 different \textsf{TimeDo} and \textsf{FreDo} models, one for each series.
Then, to compare the performances, we compare the aggregated error over all 321 models and also perform paired t-tests on the 321 models. We note that the model sizes of \textsf{TimeDo} and \textsf{FreDo} are exactly the same given the same input and output lengths.
The results of all seven datasets are shown in Tab.~\ref{tab:single}.
From the table, we can confidently conclude that learning in the frequency domain improves model performance with statistical significance.

% \subsection{Why Frequency Domain?}
% We believe there are two reasons why frequency domain is better for our models in long-term forecasting.
% First, following the notation in Fig.~\ref{fig:fredo}, we can see that the learnable modules (i.e., the black boxes) aim to predict the difference $x - \tilde{x}$.
% Due to how the baseline prediction $\tilde{x}$ is constructed, it consists more lower frequencies.
% Thus, by subtracting $\tilde{x}$ from $x$, lower frequencies are filtered out and more higher frequencies are sustained.
% This makes operating in frequency domain more favorable.
% The other reason is from the observation that spikes are quite often in the datasets, which is difficult to optimize in time domain.
% However, spikes in time domain, say the 10 in $[0, 0, 10, 0]$, will actually be transformed to a constant magnitude of 10 in frequency domain; thus making the optimization easier.
% Due to page limitation, more demonstrations and experiments are shown in Appendix~\ref{app:whyfreq}.

\vspace{-10pt}
\section{Conclusion and Limitations} \label{sec:conclusion}
\vspace{-10pt}
In this paper, we mathematically show the problem of error accumulation which can make sophisticated time series models not useful for long-term forecasting.
We propose \textsf{AverageTile}, a simple baseline model, that can achieve comparable performance to the state of the art Transformer-based model.
Building on the capability of \textsf{AverageTile}, we further propose \textsf{FreDo}, that learns in the frequency domain and greatly outperforms the state-of-the-art.
Finally, we evaluate \textsf{FreDo} and \textsf{TimeDo} univariately to validate that learning in the frequency domain is better.
The main limitation of \textsf{FreDo} is that it is designed for long-term forecasting so it does not perform better in the short-term.
Also, it is a univariate model so it does not consider interactions between series.
Thus, future research can explore multivariate frequency domain models that can also achieve better performance in the short-term, and further explore and understand the advantages of frequency domain learning.

\section*{Acknowledgements}
We thank Hilaf Hasson, Bernie Wang, and Anoop Deoras from Amazon for their helpful discussions and insights.

\bibliography{ref}
\bibliographystyle{abbrv}

\section*{Checklist}

%%% BEGIN INSTRUCTIONS %%%
% The checklist follows the references.  Please
% read the checklist guidelines carefully for information on how to answer these
% questions.  For each question, change the default \answerTODO{} to \answerYes{},
% \answerNo{}, or \answerNA{}.  You are strongly encouraged to include a {\bf
% justification to your answer}, either by referencing the appropriate section of
% your paper or providing a brief inline description.  For example:
% \begin{itemize}
%   \item Did you include the license to the code and datasets? \answerYes{See Section.}
%   \item Did you include the license to the code and datasets? \answerNo{The code and the data are proprietary.}
%   \item Did you include the license to the code and datasets? \answerNA{}
% \end{itemize}
% Please do not modify the questions and only use the provided macros for your
% answers.  Note that the Checklist section does not count towards the page
% limit.  In your paper, please delete this instructions block and only keep the
% Checklist section heading above along with the questions/answers below.
%%% END INSTRUCTIONS %%%

\begin{enumerate}

\item For all authors...
\begin{enumerate}
  \item Do the main claims made in the abstract and introduction accurately reflect the paper's contributions and scope?
    \answerYes{See last paragraph in Sec.~\ref{sec:intro}.}
  \item Did you describe the limitations of your work?
    \answerYes{See the last few sentences in Sec.~\ref{sec:conclusion}.}
  \item Did you discuss any potential negative societal impacts of your work?
    \answerNo{It is not obvious how a better forecasting model can be used in a negative way.}
  \item Have you read the ethics review guidelines and ensured that your paper conforms to them?
    \answerYes{}
\end{enumerate}

\item If you are including theoretical results...
\begin{enumerate}
  \item Did you state the full set of assumptions of all theoretical results?
    \answerYes{In Sec.~\ref{sec:erracc}.}
   \item Did you include complete proofs of all theoretical results?
    \answerYes{Not a very complex proof so all results are included.}
\end{enumerate}

\item If you ran experiments...
\begin{enumerate}
  \item Did you include the code, data, and instructions needed to reproduce the main experimental results (either in the supplemental material or as a URL)?
    \answerYes{The (anonymized) URL is in the Abstract. We also submit the code in the supplementary material.}
  \item Did you specify all the training details (e.g., data splits, hyperparameters, how they were chosen)?
      \answerYes{In Sec.~\ref{sec:datasets}, Sec.~\ref{sec:settings}, and Appendix~A.}
  \item Did you report error bars (e.g., with respect to the random seed after running experiments multiple times)?
    \answerYes{In Sec.~\ref{sec:exp}, either in the form of standard deivation or $p$-value.}
   \item Did you include the total amount of compute and the type of resources used (e.g., type of GPUs, internal cluster, or cloud provider)?
    \answerNo{Experiments do not require workstation GPUs.}
\end{enumerate}

\item If you are using existing assets (e.g., code, data, models) or curating/releasing new assets...
\begin{enumerate}
  \item If your work uses existing assets, did you cite the creators?
    \answerYes{We use public datasets and we cite them in Sec.~\ref{sec:datasets}.}
  \item Did you mention the license of the assets?
    \answerNo{We are unable to find the licenses, but we do cite the papers or URLs.}
  \item Did you include any new assets either in the supplemental material or as a URL?
    \answerYes{Our code is the new asset.}
  \item Did you discuss whether and how consent was obtained from people whose data you're using/curating?
    \answerNo{We are unable to find the information.}
  \item Did you discuss whether the data you are using/curating contains personally identifiable information or offensive content?
    \answerNo{Our data do not contain identifiable or offensive content.}
\end{enumerate}

\item If you used crowdsourcing or conducted research with human subjects...
\begin{enumerate}
  \item Did you include the full text of instructions given to participants and screenshots, if applicable?
    \answerNA{}
  \item Did you describe any potential participant risks, with links to Institutional Review Board (IRB) approvals, if applicable?
    \answerNA{}
  \item Did you include the estimated hourly wage paid to participants and the total amount spent on participant compensation?
    \answerNA{}
\end{enumerate}

\end{enumerate}

% \clearpage

\begin{appendices}

\section{Hyperparameters} \label{app:hyper}

\section{Complex-valued Model} \label{app:complex}

\section{Hyperparmeter Search in \textsf{AverageTile}} \label{app:searchbase}

\section{Model Sizes} \label{app:modelsize}

% \section{More on Why Frequency Domain?} \label{app:whyfreq}

\end{appendices}

\end{document}